\documentclass[review, 3p]{elsarticle}

\usepackage{graphicx}
\usepackage{epsfig}
\usepackage{rotating}
\usepackage{subfigure}
\usepackage{amsmath}
\usepackage{amssymb}
\usepackage{natbib}

\usepackage{hyperref}
\usepackage{url}
\usepackage{array}
\usepackage{color}

\journal{ISPRS Journal of Photogrammetry and Remote Sensing}

\begin{document}

\begin{frontmatter}



\title{Binary Patterns Encoded Convolutional Neural Networks for Texture Recognition and Remote Sensing Scene Classification}


\author{Rao~Muhammad~Anwer$^{1}$, Fahad~Shahbaz~Khan$^{2}$, Joost~van~de~Weijer$^{3}$, Matthieu~Molinier$^{1,4}$, Jorma~Laaksonen$^{1}$}

\address{ $^{1}$Department of Computer Science, Aalto University School of Science, Finland\\
 $^{2}$Computer Vision Laboratory, Link\"oping University, Sweden\\
$^{3}$Computer Vision Center, CS Dept. Universitat Autonoma de Barcelona, Spain\\
$^{4}$VTT Technical Research Centre of Finland Ltd - Remote Sensing team.\\}

\begin{abstract}

Designing discriminative powerful texture features robust to realistic imaging conditions is a challenging computer vision problem with many applications, including material recognition and analysis of satellite or aerial imagery. In the past, most texture description approaches were
based on dense orderless statistical distribution of local features. However, most recent approaches to texture recognition and remote sensing scene classification are based on Convolutional Neural Networks (CNNs). The \emph{de facto} practice when learning these CNN models is to use RGB patches as input with training performed on large amounts of labeled data (ImageNet). In this paper, we show that Local Binary Patterns (LBP) encoded CNN models, codenamed TEX-Nets, trained using mapped coded images with explicit LBP based texture information provide complementary information to the standard RGB deep models. Additionally, two deep architectures, namely early and late fusion, are investigated to combine the texture and color information. To the best of our knowledge, we are the first to investigate Binary Patterns encoded CNNs and different deep network fusion architectures for texture recognition and remote sensing scene classification. We perform comprehensive experiments on four texture recognition datasets and four remote sensing scene classification benchmarks: UC-Merced with 21 scene categories, WHU-RS19 with 19 scene classes, RSSCN7 with 7 categories and the recently introduced large scale aerial image dataset (AID) with 30 aerial scene types. We demonstrate that TEX-Nets provide complementary information to standard RGB deep model of the same network architecture. Our late fusion TEX-Net architecture \emph{always} improves the overall performance compared to the standard RGB network on both recognition problems. Furthermore, our final combination leads to consistent improvement over the state-of-the-art for remote sensing scene classification.

\end{abstract}


\end{frontmatter}




\section{Introduction}
\label{sec:introduction}

Texture analysis in real-world images, robust to variations in scale, orientation, illumination or other visual appearance, is a challenging computer vision problem with many applications, including object classification and remote sensing. Over the years, a variety of texture analysis approaches
have been proposed in literature~\cite{Ojala02b,Zhang07,Varma10j,Liu16k,Liu17j} to capture different properties of texture, such as spatial structure, roughness, contrast, regularity, and orientation in images. Most successful texture description methods are based on orderless distribution of local features leading to the development of several classification frameworks, including histograms of vector quantized filter responses~\cite{Leung96k}, textons theory~\cite{Leung01j}, bag-of-visual-words~\cite{Csurka04} and later the Fisher Vector~\cite{Perronnin07k}. In this paper, we tackle the issue of learning robust texture description for texture recognition \emph{and} remote sensing scene classification.

The first problem investigated in this paper is that of texture recognition, where the task is to associate each texture image to its respective texture category. Texture recognition plays a crucial role in many applications, related to biomedical imaging, material recognition, document image analysis, and biometrics. The problem of texture recognition can be divided into two phases: the texture description stage and the classification phase. Generally, much attention has been focused on the texture description phase since it is challenging to design powerful texture features robust to imaging conditions. One of the most successful approaches to texture description is that of Local Binary Patterns (LBP)~\cite{Ojala02b} and its variants. The standard LBP descriptor~\cite{Ojala96bb} is invariant to monotonic gray scale changes and is based on the signs of differences of neighboring pixels in an image. The LBP descriptor was later extended~\cite{Ojala02b} to obtain multi-scale, rotation invariant and uniform representations and has been successfully employed in other tasks, including object detection~\cite{Zhang11h}, face recognition~\cite{Ahonen04h}, and remote sensing scene classification~\cite{Chen16jk}.

The second problem investigated in this paper is that of
remote sensing scene classification. Remote sensing scene classification is a challenging and open research problem crucial for understanding high-resolution remote sensing imagery with numerous applications including vegetation mapping, urban planning, land resource management and environmental monitoring. In this problem, the task is to automatically associate a semantic class label to each high-resolution remote sensing image containing multiple land cover types and ground objects. The problem is challenging due to several factors, such as large intra-class variations, changes in illumination due to images extracted at different times and seasons, small inter-class dissimilarity and scale variations. Several existing approaches either rely on using low-level visual features~\cite{Santos10j,Yang13jk,Chen16jk}, such as color, shape or using combination of visual features~\cite{Bin13jj,Chen15jk}. Contrary to approaches based on low-level visual features, mid-level remote sensing scene classification methods tackle the problem by encoding low-level features into a holistic high-order statistical image representation. Popular mid-level approaches include bag-of-words (BOW) variants~\cite{Chen11jj,Yang10j}, spatial extensions to BOW~\cite{Yang11j,Shizhijk}, semantic BOW using topic models~\cite{Kusumaningrum14k,Yanfei15k}, and unsupervised feature learning~\cite{Zhang15jj,Fan15jj}.

Recently, Convolutional Neural Networks (CNNs) have revolutionised computer vision, being the catalyst to significant performance gains in many vision applications, including texture recognition~\cite{Cimpoi16j} and remote sensing scene classification~\cite{Fan15jk,Penatti15j}. CNNs and other "deep networks" are generally trained on large amounts of labeled training data (e.g. ImageNet~\cite{Deng09a}) with raw image pixels with a fixed size as input. Deep networks consists of several convolution and pooling operations followed by one or more fully connected (FC) layers. Several works~\cite{Azizpour14h,Oquab14k} have shown that intermediate activations of the FC layers in a deep network, pre-trained on the ImageNet dataset, are general-purpose features applicable to visual recognition tasks. Deep features based approaches have shown to provide the best results in recent evaluations for texture recognition~\cite{Liu16k} and remote sensing scene classification~\cite{Xia177}.

As mentioned above, the \emph{de facto} practice is to train deep models on the ImageNet dataset using RGB values of the image patch as an input to the network. These pre-trained RGB deep networks are typically employed in state-of-the-art methods for texture recognition and remote sensing scene classification. Interestingly, in a recent performance evaluation for texture recognition~\cite{Liu16k}, the hand-crafted LBP texture descriptor and its variants were shown to provide competitive performance compared to deep features based methods especially in the presence of rotations and several types of noise. In addition to texture recognition, LBP and its variants have been successfully employed for remote sensing scene classification~\cite{Santos10j,Chen16jk}. Moreover, the work of~\cite{Levi15c} proposes to train CNNs on pre-processed texture coded images in addition to RGB for emotion recognition. Motivated by these observations, we investigate the impact of integrating LBP within deep learning architectures for texture recognition and remote sensing scene classification.

The combination of multiple feature streams into a single architecture has recently been a subject of intense study. It is being investigated in the context of action recognition~\cite{Simonyan14k,Cheron15k,Feichtenhofer16j}, RGB-D~\cite{Hoffman16k}, and multi-modal networks~\cite{Reed16k,Akira16k}. In the aforementioned multiple feature streams action recognition approaches, the spatial stream captures the appearance information by using RGB images as input to the network and the temporal stream captures the motion information by using dense optical flow images as input to the network. The spatial and motion streams are then fused since they contain complimentary information. Inspired by the success of these two-stream deep networks, we propose a two-stream deep architecture where texture coded mapped images are used as the second stream and fuse it with the normal RGB image stream. The two network streams can be fused at different stages in the deep architecture. In the first strategy, termed as late fusion, the RGB and texture streams are trained separately and combined at a later stage by fusing them at the FC layers. In the second strategy, termed as early fusion, the two streams are joined at an early stage by aggregating the RGB and texture coded image channels as an input, in order to train a joint two-stream deep model. To the best of our knowledge, we are the first to investigate these two fusion strategies, to combine RGB and texture streams, in the context of texture recognition and remote sensing scene classification.

\noindent\textbf{Contributions:} In this work we investigate the problem of learning robust texture description by integrating one of the most popular hand-crafted texture descriptor, Local Binary Patterns (LBP), within deep learning architectures for texture recognition and remote sensing scene classification. To this end, we propose deep models, which we call TEX-Nets, by designing a two-stream deep architecture where texture coded mapped images are used as the second stream and fuse it with the normal RGB image stream. To obtain the texture coded mapped images, we first extract LBP based codes from an image.
Afterwards, as in~\cite{Levi15c}, the unordered LBP code values are mapped to points in a 3D metric space. The mapping is performed by employing Multi Dimensional Scaling (MDS) using code-to-code dissimilarity scores based on approximated Earth Mover's Distance (EMD). We further evaluate two fusion strategies, early and late fusion, to combine RGB and texture streams for texture recognition and remote sensing scene classification.

The proposed approach is first evaluated on a selection of texture benchmark datasets to demonstrate the overall effectiveness of the approach, and then applied to several remote sensing benchmark datasets to demonstrate its potential and applicability to remote sensing scene classification. The results of our experiments suggest that our late fusion TEX-Net architecture provides superior results compared to the early fusion TEX-Net architecture. Further, the proposed late fusion TEX-Net architecture \emph{always} improves the overall performance compared to the standard RGB stream deep network architecture. Lastly, our final combination leads to performance superior to the state-of-the-art without employing fine-tuning or ensemble of RGB network architectures, for remote sensing scene classification.

\section{Related Work}
\label{sec:relatedwork}
Here, we briefly review the Local Binary Patterns (LBP) and its variants, deep learning and state-of-the-art in texture recognition and remote sensing scene classification.

\noindent\textbf{Local Binary Patterns:} In the field of texture recognition, local binary patterns (LBP) ~\cite{Ojala02b} is one of the most commonly used texture description approaches. Besides texture recognition, LBP based texture description has been applied to other vision tasks, including face recognition~\cite{Tan07h}, gender recognition~\cite{khan14gg}
and person detection~\cite{Wang09k}. The LBP descriptor works by thresholding intensity values of a pixel around its neighborhood. The threshold is computed from the intensity of each neighborhood's center pixel. A circular symmetric neighborhood is employed by interpolating the locations not exactly at the center of a pixel. A variety of LBP variants have been proposed in literature, including Local Ternary Patterns~\cite{Tan10j}, Local Binary Pattern Variance~\cite{Zhenhua10j}, Noise Tolerant Local Binary Patterns~\cite{Fathi12j}, Completed Local Binary Patterns~\cite{Zhenhua10jj}, Extended Local Binary Patterns~\cite{Liu12jj} and Rotation Invariant Local Phase Quantization~\cite{Ojansivu09k}. In addition to the introduction of different LBP variants, the fusion of LBP descriptor with color features have also been investigated in previous studies~\cite{Maenpaa04p,khan15j}.

\noindent\textbf{Deep Learning:} In recent years, Convolutional Neural Networks (CNNs)~\cite{LeCun89k} have shown to provide excellent performance for many computer vision tasks. CNNs are generally trained using large amount of labeled training samples and take fixed sized RGB
images as input to a series of convolution, normalization and pooling operations (termed as layers). The network typically ends with several fully-connected (FC) layers, used to extract features for recognition. Several attempts have been made to improve deep network architectures, including increasing the depth of the network by introducing additional convolutional layers~\cite{Simonyan15k,Kaiming16k}. In addition to RGB based appearance networks, other modalities such as motion and depth have also been used to construct multi-cue deep networks for action recognition~\cite{Simonyan14k} and RGB-D object recognition~\cite{Eitel15k}.

\noindent\textbf{Deep Learning for Remote Sensing Image Analysis:} In recent years, deep learning methods have made a breakthrough for satellite image analysis, with several works published in the major remote sensing
journals. The most notable applications of deep neural networks
(DNNs) in remote sensing include land cover classification
with optical images~\cite{Xueyun14jjk,Adriana16k,Molinier07k},
hyperspectral image analysis~\cite{Yushi14jj,Yushi15jj,Tuia15jj}
or Synthetic Aperture Radar (SAR) image analysis~\cite{Geng15jjk}.

A large majority of published works use DNNs trained on
patches extracted from satellite images. DNNs are usually
not trained on databases of full sized satellite images (1 to
several GB per image) due to memory limitations, even on
powerful GPU servers. CNNs
are the most commonly used deep learning architectures for
the classification of optical~\cite{Xueyun14jjk} and SAR~\cite{Geng15jjk} satellite images. Because large datasets of satellite images with high quality
labels are not easily available, most of the earlier works
utilized pre-trained DNNs that were trained on computer vision
benchmark datasets (ImageNet), not on satellite images~\cite{Marmanis16jjk}.

\noindent\textbf{Texture Recognition:} A variety of texture recognition approaches have been proposed in literature~\cite{Liu16k,Pietikainen16g}. The work of~\cite{Varma10j} proposes a statistical approach to model textures based on the joint probability distribution of filter responses. The work of~\cite{Chen10j} proposes an approach based on Weber's law which consists of two components: differential excitation and orientation. An image is represented by the concatenation of these two components in a single representation. The work of~\cite{Hussain12h} introduces an approach that uses lookup-table based vector quantization for texture description. A set of low and mid-level perceptually inspired visual features are introduced by~\cite{Sharan13j} for texture recognition. A multi-resolution framework based on LBP is proposed by~\cite{Ojala02b} for rotation invariant texture recognition. As discussed earlier, LBP is one of the most successful approaches for texture recognition with several variants existing in literature~\cite{Guo10b,Ylioinas13h,Ylioinas12h}.

Other than LBP and its variants, bag-of-words based representations employing SIFT features and Fisher Vector encoding scheme have shown promising results for texture recognition~\cite{Cimpoi14k}. Recently, deep features have also been investigated for texture recognition. Bruna and Mallat~\cite{Bruna13jj} introduce the wavelet convolutional scattering network (ScatNet), where no learning is required and convolutional filters are defined as wavelets. The work of~\cite{Chan14jj} proposes a deep network based on multistage principal component analysis (PCANet). The work of~\cite{Cimpoi16j} proposes to use the convolutional layers of the deep networks as dense local descriptors encoded with Fisher Vector to obtain the final image representation.

\noindent\textbf{Our Approach:} As discussed above, most existing hand-crafted approaches employ LBP and its variants for texture description. On the other hand, deep learning based approaches have shown promising results for texture recognition and remote sensing scene classification. Despite the success of deep features, the hand-crafted LBP texture descriptor and its variants
have been shown to provide competitive performance
compared to deep feature based methods especially in
the presence of rotations and several types of noises in a recent performance evaluation for texture recognition~\cite{Liu16k}. Moreover, the deep features based texture recognition and remote sensing scene classification methods employ deep networks pre-trained on the ImageNet dataset using \emph{RGB images} as input. This motivates us to investigate the impact of
integrating texture features, in particular, the popular hand-crafted LBP texture
descriptor within deep learning architectures. We investigate fusion strategies by constructing a two-stream deep architecture where texture coded mapped
images are used as the second stream and fuse it with
the normal RGB image stream. To the best of our knowledge, we are the first to investigate the two fusion
strategies in a two-stream deep architecture, to combine RGB and texture streams, in the context of texture recognition and remote sensing
scene classification. This paper is an extended version of our earlier work~\cite{RaoICMR17}. We have extended our experiments by evaluating the proposed approach for remote sensing scene classification application with results on four challenging benchmarks. In addition, we also provide an analysis of our two-stream deep architecture on the ImageNet dataset.

\section{Binary Patterns Encoded Convolutional Neural Networks}
\label{sec:method}

Here, we first describe the construction of deep models based on texture coded
mapped images. Afterwards, we investigate different strategies
to fuse the texture coded mapped stream with the normal RGB image stream.

\subsection{Mapped LBP Codes}
\label{sec:Mapped_Codes}

As discussed earlier, Local Binary Patterns (LBP) has shown competitive performance for texture recognition and is one of the most commonly employed approaches for texture description. LBP features describe the neighborhood of a pixel by its binary derivatives. These
binary derivatives are then used to form a short code to describe the neighborhood of the
pixel. The short LBP codes are binary numbers (lower than threshold (0) or higher than the
threshold (1)). Each LBP code can be considered as a micro-texton since each pixel is assigned a code of the texture primitive with its best local neighborhood match. Several local
primitives are detected by the LBP operator, including flat areas, edges, corners, curves, and edge ends.
The primitive version of LBP operator considered only the eight-neighbors of a pixel, while using the center pixel value as a threshold. Later variants extended the primitive LBP operator to consider all circular neighborhoods with any number of pixels. Given an image $f(a_{c},b_{c})$ of size $H\times W$, with $(a_{c}\in \{0, ..., H-1\}, b_{c}\in \{0, ..., W-1\})$. Here, $(a_{c},b_{c})$ are the coordinates of the center pixel of a circular local neighborhood $(P,R)$, where $P$ denotes the number of sampling points and $R (R>0)$ is the
circle radius of the of local neighborhood. The LBP code (a $P$-bit word) describing the local image texture around the the center pixel is computed as,

\begin{equation}
LBPC_{P,R}(a_{c},b_{c}) = \sum\limits_{p=0}^{{P-1}} s(f(a_{p},b_{p})-f(a_{c},b_{c}))2^{p},
 \label{lbp_main_eq}
\end{equation}

where the thresholding function $s(t)$ is defined as:

\begin{equation}
s \left( t \right) =\left\{ {\begin{array}{*{20}{c}}
   {\;\;0\;\;{\rm{for}}\;t < 0\;\;\;}  \\
   {1\;\;{\rm{for}}\;t \geq 0}.  \\
\end{array}} \right.
\end{equation}

The standard LBP computation results in $2^{p}$ distinct values for the LBP code. In case of an 8 pixel neighborhood, the LBP code
computation results in a binary string of eight-bit numbers between 0 and 255. The
final image representation is obtained by computing the histogram
as a distribution of LBP codes over an entire image region. The resulting feature vector normalizes for
translation and is
invariant to monotonic changes in the gray scale.

As discussed above, the LBP codes are generally pooled as histogram representations and employed as an input to a discriminative classifier, such as Support Vector Machines (SVMs). Instead, due to the overwhelming recent success of deep learning, it is worth investigating to integrate the strength of LBP descriptor within the CNN architectures. A straightforward integration strategy is to train deep models by directly using LBP codes as CNN inputs. However, such a strategy is not applicable since the convolution operations, equivalent to a weighted average of the input values, performed within CNN models are unsuitable for the unordered nature of the LBP code values.

Recently, the work of~\cite{Levi15c} provides a solution to this problem within the context of texture description for emotion recognition. They propose to map the LBP codes to points in a 3D metric space in which the Euclidian distance approximates the distance between the LBP codes. After the transformation of the LBP codes they can be averaged together using convolution operations within CNN models.

\begin{figure}[t]

\centerline{\includegraphics[width=8.7cm]{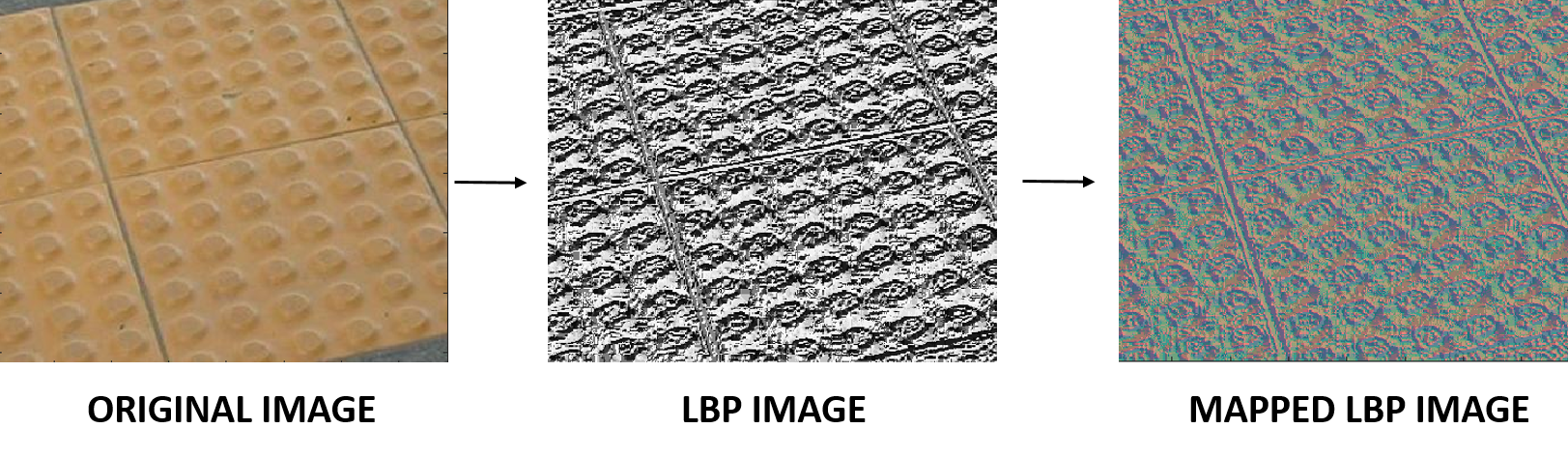}}

\caption{Example of the texture coded mapped image (visualized here in color). The mapped LBP image is obtained by converting LBP codes (shown as grayscale values) into a 3D metric space.}
\label{fig:mapped_image}
\end{figure}

The method is based on defining a distance $\delta_{j,k}$ between the LBP codes $LBPC_j$ and $LBPC_k$. The authors of~\cite{Levi15c} choose the Earth Movers Distance (EMD)~\cite{Rubner0jj} because it accounts for both the different bit values and their locations. Having defined the distance between the LBP codes, it is now possible to look for a mapping of the LBP codes into a $D$-dimensional space which approximately preserves this distance. This mapping can be found by applying Multi Dimensional Scaling (MDS)~\cite{BorgGroenen2005}, such that:

\begin{equation}
\delta_{j,k} \approx \parallel L_{j} - L_{k} \parallel = \parallel MDS(LBPC_{j}) - MDS(LBPC_{k}) \parallel.
 \label{MDS_main_eq}
\end{equation}

where $L_j=MDS(LBPC_j)$ is the mapping of code $j$ into the $D$-dimensional space. Applying this mapping allows us to transfer the LBP codes into a representation which can be used as input to a CNN. In~\cite{Levi15c} they experimented with the optimal dimensionality $D$ and found that good results were obtained with $D=3$. In this work, we use the same settings and in addition also investigate an early fusion scheme with $D=1$. We refer to~\cite{Levi15c} for more details. Figure~\ref{fig:mapped_image} shows an example image converted to LBP codes (middle). The LBP codes are mapped to a 3D metric space (right) and normalized before used as an input to CNNs.

\subsection{Texture Coded Two-Stream Deep Architecture}
\label{sec:two-stream}
As described earlier, the de-facto standard when training deep models is to use raw RGB pixels values of an image as input. These RGB based deep networks have achieved state-of-the-art results for texture recognition recently~\cite{Cimpoi16j} and remote sensing scene classification~\cite{Fan15jk,Penatti15j}. In this work, we investigate to what extent texture coded deep networks complement the standard RGB based CNN models in two classification problems: texture recognition and remote sensing scene classification. To this end, we design a two-stream deep architecture, referred as TEX-Nets, using both texture coded
mapped images (section~\ref{sec:Mapped_Codes}) and raw RGB pixel values. Our TEX-Nets models are trained on the ImageNet ILSVRC-2012 dataset~\cite{Deng09a}. We employ two different architectures to validate our approach: the VGG-M architecture~\cite{Chatfield14h} which is similar to Zeiler and Fergus network~\cite{Zeiler14c} and the ResNet architecture~\cite{Kaimingcvpr15}. The VGG-M network comprises of five convolutional and three fully-connected (FC) layers. The VGG-M network takes as input an image of $224\times224$ pixels. The first convolutional layer employs smaller stride (1) and receptive field (or the filter size). The second convolutional layer uses a relatively larger stride (2 compared to 1). The number of convolution filters is 96 in the first convolutional layer, 256 in the second convolutional layer and 512 in the third and last convolutional layers. During training, the learning rate is set to $0.001$, a weight decay that acts as a regularizer and helps reducing the training error of the model is set to $0.0005$. The momentum rate is associated with the gradient descent algorithm used to minimize the objective function and is set to $0.9$. We also employ the ResNet-50 architecture~\cite{Kaimingcvpr15} which is a 50 layer Residual Network. This architecture is based on residual learning framework that facilitates efficient training of deeper networks by reformulating the layers as learning residual functions with reference to the layer inputs. The ResNet-50 architecture takes as input an image of $224\times224$ pixels. For the first 30 training iterations, the learning rate is set to $0.1$. For the second and the last 30 training iterations, the learning rate is set to $0.01$ and $0.001$ respectively. The momentum and the weight decay is set to $0.9$ and $0.0001$ respectively.

Next, we investigate strategies to fuse the two network streams at different stages in the deep architectures.

\noindent \textbf{Late Fusion:} In this strategy, both standard (RGB) and texture coded network streams are trained separately on the ImageNet dataset. The standard RGB network stream takes RGB values as input, whereas the second network stream takes texture coded mapped
images as input. These texture coded mapped images are obtained by first employing the LBP encoding that converts intensity values in an image to one of the 256 LBP code values. The LBP code values are then mapped into a 3D metric space (section~\ref{sec:Mapped_Codes}). The resulting 3-channel texture coded mapped
images are then used as input to CNN models. Despite being efficient to compute, the texture coded mapped
images still introduce a bottleneck if done on-the-fly. We therefore pre-compute these texture coded mapped images before training the deep network. Once separately trained, the RGB and texture coded network streams are combined at a later stage by fusing them in the FC layers in the VGG-M architecture. In case of ResNet architecture, late fusion is performed before the softmax loss. The two-stream late fusion strategy has been previously used in action recognition to combine spatial (RGB) and temporal (flow) information~\cite{Simonyan14k,Cheron15k}.

\begin{figure}[t]
\centerline{\includegraphics[width=8.7cm]{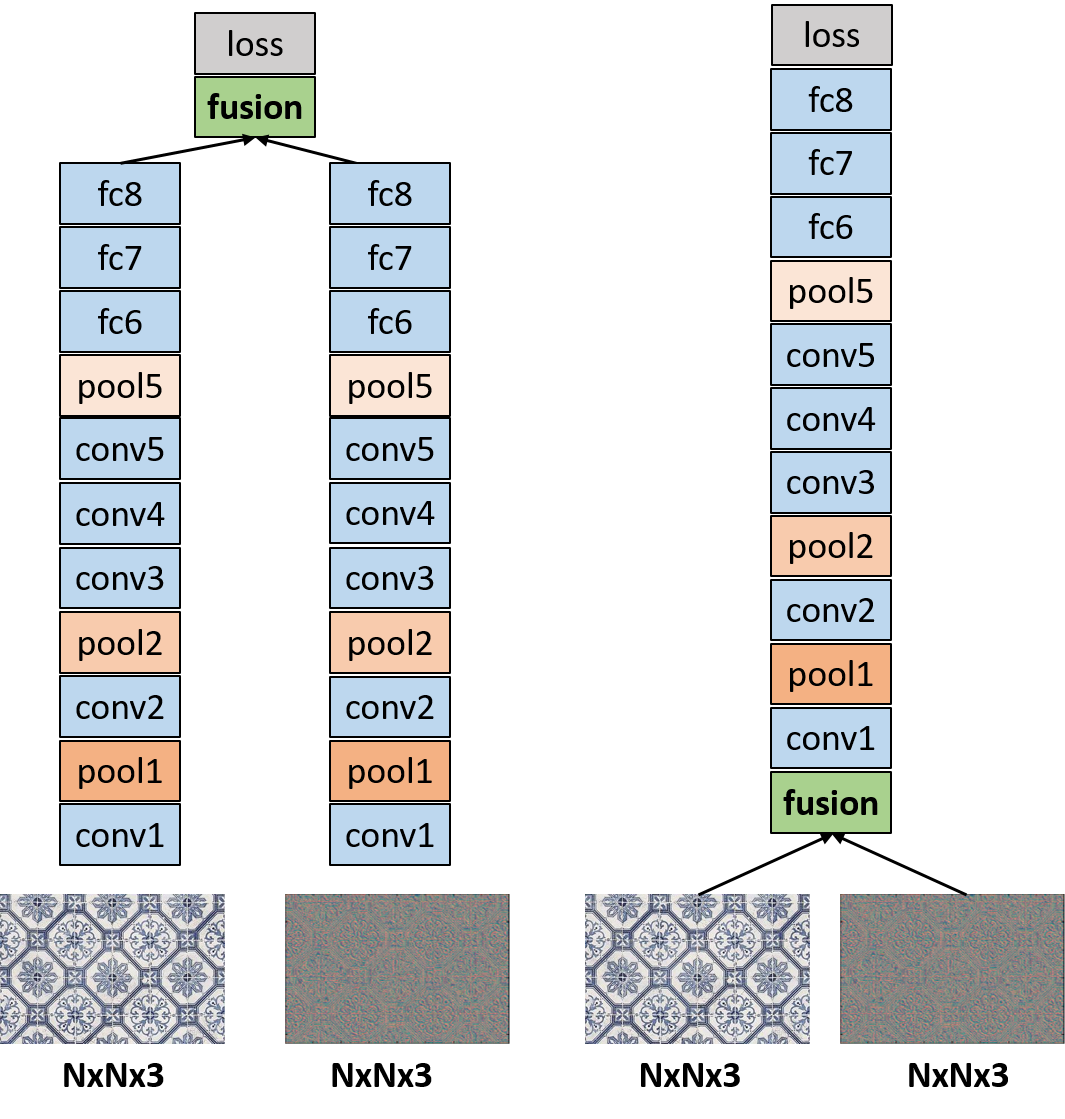}}
\caption{Two-stream deep fusion VGG-M architectures. The left example shows late fusion architecture where the deep models trained using RGB and texture coded mapped images are kept separately. The point of fusion, to combine the two network towers, is in the FC layer. The right example shows early fusion architecture where the point of fusion is the input to the network. As a result, a joint deep model is trained by aggregating the RGB and texture image channels as an input to the network.}
\label{fig:fusion_ef_lf}
\end{figure}

\noindent \textbf{Early Fusion:} Other than late fusion, we also investigate an alternative strategy, termed as early fusion, where the point of fusion is the input to the network. In the early fusion based two-stream network architecture, a joint deep model is trained by aggregating the RGB and texture coded mapped image channels as an input to the deep network. As a result, the input to CNN is an image of $224\times224\times6$ dimensions. We employ same early fusion strategy for both VGG-M and ResNet architectures. We also  investigated converting the 3-channel mapped coded images into a single channel and combining it with the three RGB channels. In both networks, the filters are learned jointly on the RGB and texture coded images. Figure~\ref{fig:fusion_ef_lf} shows both early and late fusion based two-stream deep fusion VGG-M architectures designed to combine the color and texture image streams.

\begin{table}[t!]
\centering \scriptsize  \addtolength{\tabcolsep}{0.1pt}

\begin{tabular}{|l|c|c|c|c|c|}
\hline Method & Architecture & Channels & Top-1 Error (\%) & Top-5 Error (\%)  \\
\hline\hline
Standard RGB (Baseline) & VGG-M & 3 & 37.6   & 15.9  \\
TEX-Net Standard  & VGG-M & 3 & 45.8  & 21.9  \\
TEX-Net-EF-6ch  & VGG-M & 6 & 39.3   & 17.7 \\
TEX-Net-EF-4ch  & VGG-M & 4&  37.1   & 15.5  \\
TEX-Net-LF  & VGG-M & 6 & 34.4   & 13.8  \\
\hline\hline
Standard RGB (Baseline) & ResNet & 3 & 25.4   & 8.0  \\
TEX-Net-LF  & ResNet & 6 & \textbf{23.7}   & \textbf{7.0}  \\
\hline
\end{tabular}

\caption{
  Classification performance comparison of our two-stream
deep TEX-Net architectures with the standard RGB network on the ImageNet ILSVRC 2012 validation data. In case of VGG-M architecture, we show comparison with both early and late fusion TEX-Net models: early fusion architecture aggregating the RGB and 3 mapped coded channels (TEX-Net-EF-6ch), early fusion architecture aggregating the RGB and a single mapped coded channel (TEX-Net-EF-4ch) and the late fusion architecture (TEX-Net-LF) combining separate streams of RGB and texture networks. We also show results based on only mapped coded images (TEX-Net Standard), without color information. In case of ResNet architecture, we show the comparison between our late fusion approach and the standard RGB network.
}
\label{ImageNet_comp_tab}
\end{table}

\begin{figure*}[t]
\centerline{\includegraphics[width=17.0cm]{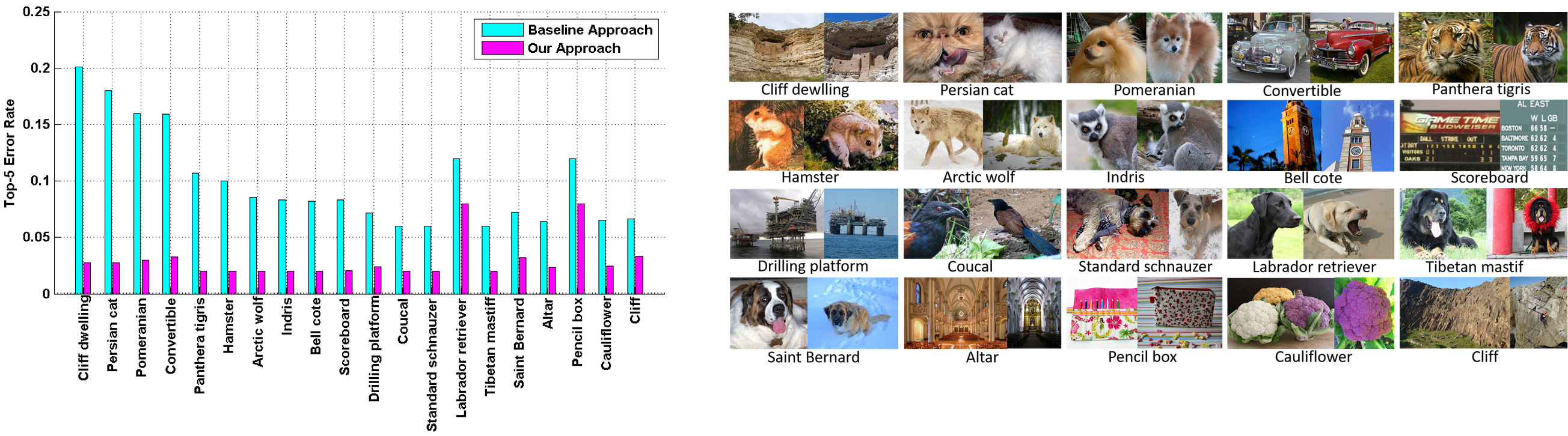}}
\caption{Object categories in the ImageNet dataset where our late fusion two-stream deep architecture provides significant reduction in the top-5 error compared to the baseline standard RGB deep network. On the left, we show the comparison (in top-5 error) and on the right, we show example images from these object categories (left to right). Both approaches are based on VGG-M architecture.}
\label{fig:per_class_imagenet}
\end{figure*}

\noindent\textbf{Training TEX-Nets on ImageNet:} As described earlier, we train our TEX-Nets from scratch on the ImageNet 2012 dataset employed in ImageNet Large-Scale Visual Recognition Challenge (ILSVRC). The dataset consists of 1000 object classes and 1.2 million training images, 50,000 validation images, and 100,000 test images. On this dataset, the results are measured by top-1 and top-5 error rates. The error rates are computed from the predictions using the deep network and obtaining the predicted class multinomial distributions. The top-5 error is the fraction of test images for which the true label is not among the five labels (the 5 predictions with the highest probabilities) considered most probable by the deep model. The top-1 error is computed by evaluating if the top class (the one having the highest confidence) is the same as the correct (target) label. Table~\ref{ImageNet_comp_tab} shows the classification performance comparison, based on VGG-M architecture, of our early and late fusion based two-stream deep TEX-Net architectures with the standard RGB deep network on the ILSVRC 2012 dataset. The standard baseline RGB network achieves top-1 and top-5 errors of $37.6\%$ and $15.9\%$ respectively. Our late fusion deep architecture significantly reduces the error with an absolute reduction of $3.2\%$ in the top-5 error, compared to the standard RGB network. The late fusion architecture results in increasing the number of network parameters by a factor of 1.4, compared to the standard RGB. We therefore also train a six channel early fusion network by increasing the network depth with a factor of 1.4, resulting in same number of parameters as late fusion. This improves the results for the six channel early fusion architecture. However, it still provides inferior results (35.3 top-1 error) compared to the late fusion architecture (34.4 top-1 error).

We further validate the effectiveness of our late fusion two-stream approach by employing the ResNet-50 architecture. Table~\ref{ImageNet_comp_tab} shows the classification performance comparison of our approach and the standard RGB network. Both networks are trained from scratch on the ImageNet dataset.  The standard baseline RGB network achieves top-1 and top-5 errors of $25.4\%$  and $8.0\%$ respectively. Our late fusion based TEX-Net ResNet architecture (TEX-Net-LF) reduces the error with top-1 and top-5 errors of $23.7\%$ and $7.0\%$ respectively.

Figure~\ref{fig:per_class_imagenet} shows 20 object categories from the ImageNet dataset where our late fusion two-stream deep architecture provides the largest reduction in the top-5 error rate, compared to the standard RGB deep network. For majority of the depicted classes it is likely that a good texture representation is crucial for correct classification. Consequently, the aforementioned results suggest that our late fusion two-stream deep architecture provides superior results compared to both standard RGB and early fusion.

\section{Experimental Results}
\label{sec:exp_section}
Here, we start by evaluating our TEX-Net deep models for the texture recognition problem.
We then provide a comparison of our approach with the standard RGB based deep network in the remote sensing scene classification task. Finally, we compare the performance of our approach with state-of-the-art remote sensing scene classification results reported in literature.

\begin{figure}[t!]
\begin{center}
\includegraphics[width=8.5cm]{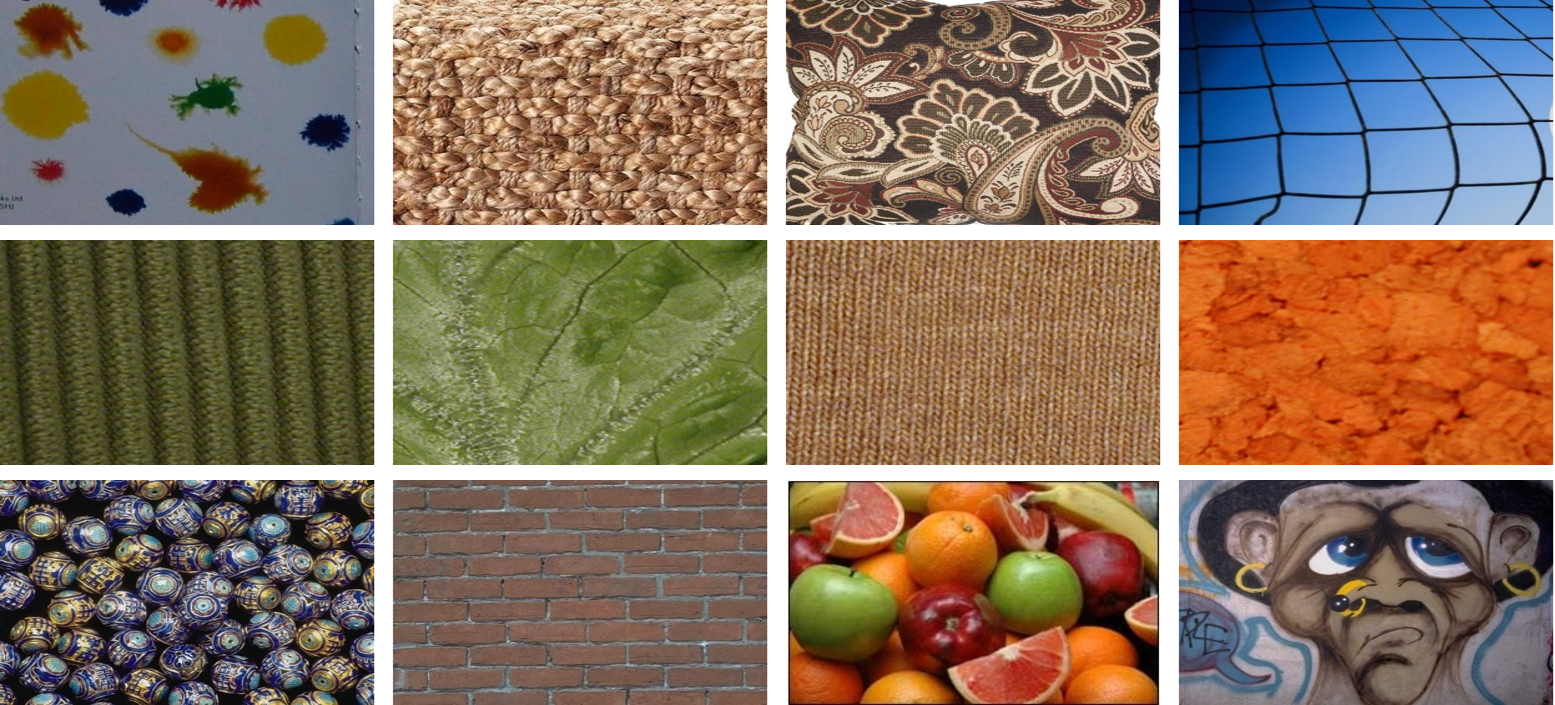}

\end{center}
\caption{Example images from the four texture datasets: DTD, KTH-TIPS-2a, KTH-TIPS-2b and Texture-10.}
\label{dataset_texture_fig1}

\end{figure}

\subsection{Texture Recognition}
 \label{sec:texture_exp_subsection}

 We evaluate our approach by performing experiments on four challenging texture datasets: DTD, KTH-TIPS-2a, KTH-TIPS-2b and Texture-10. Figure~\ref{dataset_texture_fig1} shows example images from the four texture datasets.

\noindent \textbf{DTD:} The DTD dataset consists of 5640 images from 47 texture classes, collected from the web. Each texture class consists of 120 images with the dataset equally divided into training, validation and test. The training and test splits are provided by the authors.

\noindent \textbf{KTH-TIPS-2a:} The KTH-TIPS-2a dataset consists of 11 texture classes. The 4752 images are captured at 9 different scales, 3 poses and 4 different illumination conditions. Similar to previous works~\cite{Chen10j,Caputo05h,Sharma12h}, average classification performance
is reported over the 4 test runs. In each run, images from 1 sample are used for testing while the images from the remaining 3 samples are used as a training set.

\noindent \textbf{KTH-TIPS-2b:} The KTH-TIPS-2b dataset consists of 11 texture categories. Here, images from 1 sample are used for training while all the images
from remaining 3 samples are used for testing in each test run.

\noindent \textbf{Texture-10:} The Texture-10 dataset consists of 400 images of 10 different texture categories. For each texture category, 25 images are used for training and 15 images are used for testing.

\noindent \textbf{Experimental Setup:} As discussed earlier, both the TEX-Net networks and the standard RGB deep network are trained from scratch on ImageNet 2012 training set. The deep models are trained by employing the Matconvnet library~\cite{Vedaldi15k}. We evaluate our VGG-M architecture based deep models, pre-trained on ImageNet, as feature extractors on texture datasets. We therefore remove the last fully-connected layer (FC8) of the VGG-M networks which performs 1000-way ImageNet (ILSVRC) classification, and instead use 4096 dimensional activations from the FC7 (second last) layer as image features. The resulting image features are $\mathrm{L_{2}}$-normalised and input to a linear SVM classifier. Throughout our experiments, we fixed the weights (no fine-tuning) of all the pre-trained deep VGG-M networks for fair comparison. In all cases (datasets), the results are reported as average recognition accuracy over all texture categories in a texture dataset. The classification is performed by employing one-versus-all SVMs with linear kernel. The category label from the classifier providing the highest confidence is assigned to the test instance. The overall classification results are then obtained by calculating the average of the classification scores of all texture classes in a dataset.

In case of of ResNet architecture, we fine-tuned both the standard RGB and our late fusion two-stream model to perform classification in an end-to-end fashion. For fine-tuning on each dataset, we use the training samples with a batch size of 80 and a momentum value of $0.9$. The learning rate is set to $0.005$.

\subsection{Baseline Comparison}
We compare our TEX-Net deep models with the standard RGB based CNN approach to validate whether RGB and texture coded mapped images contain complementary information. We further evaluate both early and late fusion two-stream deep architectures (section~\ref{sec:two-stream}) for combining texture and color information. For fair comparison, we use the same network architecture (VGG-M) together with the same set of parameters for all the deep models. Table~\ref{baseline_comp_texture_tab} shows the baseline comparison on four texture datasets. In case of VGG-M architecture, the standard RGB deep network provides a mean accuracy of $63.4\%$ on the DTD dataset. The two early fusion based two-stream deep architectures (TEX-Net-EF-6ch and TEX-Net-EF-4ch) slightly improve the accuracy over the standard RGB, with mean classification scores of $64.0\%$ and $64.6\%$ respectively. The image representation based on the TEX-Net standard model provides a classification score of $55.9\%$. On this dataset, the best results are obtained with our late fusion based two-stream deep ResNet architecture. On the KTH-TIPS-2a dataset, the standard RGB deep network provides a mean classification rate of $81.8\%$. Our TEX-Net standard model based on texture coded mapped images provides a classification score of $68.6\%$. The two early fusion based two-stream deep architectures (TEX-Net-EF-6ch and TEX-Net-EF-4ch) provide slight improvement in performance over standard RGB, with mean recognition scores of $82.6\%$ and $83.4\%$ respectively. Our late fusion based two-stream deep architecture achieves a mean classification rate of $85.3\%$. When using the ResNet architecture, our late fusion approach provides superior results compared to the standard RGB network.

\begin{table}[t!]
\begin{center}
\scriptsize\addtolength{\tabcolsep}{-1.75pt}
\begin{tabular}{|c|c|c|c|c|c|c|c|c|c|}
    \hline
    &{ \textbf{Architecture}}& { \textbf{DTD}}&{\textbf{KTH-TIPS-2a}}&{\textbf{KTH-TIPS-2b}}&{\textbf{Texture-10}}\\\hline

   \hline Standard RGB & VGG-M & 63.4 $\pm 0.7$ & 81.8 $\pm 5.1$ & 72.9 $\pm 2.1$&87.3\\

   \hline TEX-Net Standard & VGG-M & 55.9 $\pm 1.1$& 68.6 $\pm 5.3$ & 60.2 $\pm 2.9$&81.7\\
   \hline TEX-Net-EF-6ch & VGG-M & 64.0 $\pm 0.8$& 82.6 $\pm 5.5$ & 73.6 $\pm 2.6$&89.1\\
   \hline TEX-Net-EF-4ch& VGG-M & 64.6 $\pm 0.9$& 83.4 $\pm 5.3$ & 73.8 $\pm 2.7$ &89.3\\
   \hline TEX-Net-LF & VGG-M & 68.2 $\pm 0.8$& 85.3 $\pm 5.6$ & 75.5 $\pm 2.7$&91.3\\
     \hline   \hline
      Standard RGB & ResNet & 69.6 $\pm 0.7$& 83.3 $\pm 5.1$ & 75.2 $\pm 2.9$&90.1\\
     \hline TEX-Net-LF & ResNet & \textbf{73.6} $\pm \textbf{0.6}$& \textbf{88.3} $\pm \textbf{5.3}$ & \textbf{78.0} $\pm \textbf{2.8}$&\textbf{92.3}\\
    \hline
\end{tabular}
\end{center}
\caption{%
  Comparison (in $\%$) of our approaches with the standard RGB deep network on four texture datasets. In case of VGG-M architecture, we show comparison with our different TEX-Net models: based on only mapped coded images (TEX-Net Standard), early fusion two-stream architectures combining either the RGB and 3 mapped coded channels (TEX-Net-EF-6ch) or RGB and a single mapped coded channel (TEX-Net-EF-6ch), and the late fusion architecture (TEX-Net-LF) combining standard RGB and TEX-Net standard networks. In case of ResNet architecture, we show the comparison
between our late fusion approach and the standard RGB network. For both VGG-M and ResNet architectures, our late fusion approach \emph{always} outperforms the corresponding baseline standard RGB network.}

\label{baseline_comp_texture_tab}
\end{table}

\begin{figure*}[t]
\centerline{\includegraphics[width=16.7cm]{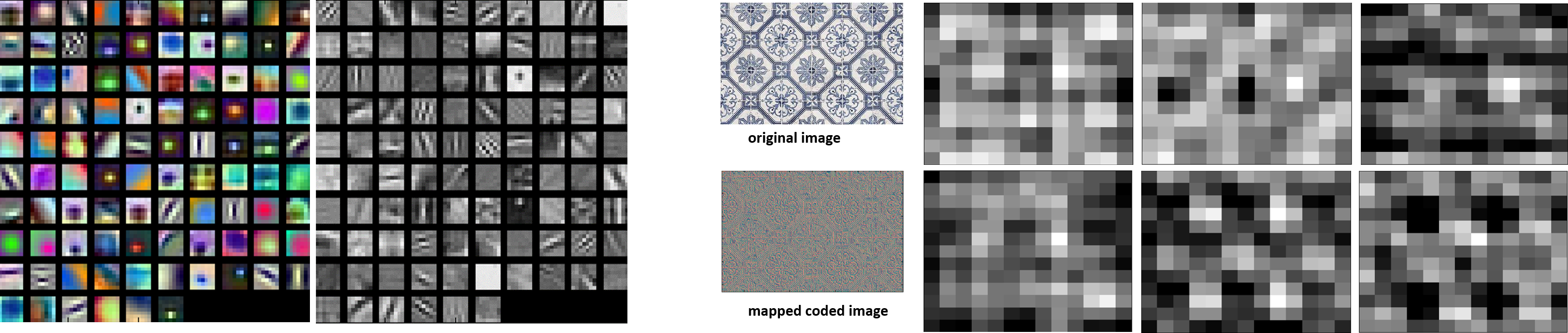}}
\caption{On the left, visualization of filter weights from the RGB and TEX-Net VGG-M model with mapped coded texture information respectively. On the right, visualization of activations with highest energy from the conv3 layer of RGB (top row) and TEX-Net (bottom row) networks on an example texture image. The TEX-Net model is trained on the texture coded mapped images (visualized here in color), obtained by converting LBP codes into a 3D metric space. In both cases, the models are based on VGG-M architecture. }
\label{activation_texture_fig1}
\end{figure*}

In case of VGG-M architecture, the baseline RGB deep network provides a mean accuracy of $72.9\%$ on the KTH-TIPS-2b dataset. The two early fusion based two-stream deep architectures (TEX-Net-EF-6ch and TEX-Net-EF-4ch) achieve mean classification scores of $73.6\%$ and $73.8\%$ respectively. When using the ResNet architecture, our late fusion based deep network provides a gain of $2.8\%$ over the standard RGB network. Finally, on the Texture-10 dataset, the standard RGB deep network achieves a mean classification score of $87.3\%$ with VGG-M architecture. Our late fusion based two-stream deep VGG-M architecture obtains a mean accuracy of $91.3\%$, leading to a gain of $4.0\%$ compared to the standard RGB VGG-M network. The best results are obtained using our late fusion approach with ResNet architecture. Figure~\ref{activation_texture_fig1} shows a VGG-M architecture based visualization of filter weights (on the left) from the RGB and TEX-Net model respectively and a visualization of activations (on the right) with the highest energy from the conv3 layer of the RGB (top row) and
TEX-Net (bottom row) networks on an example texture image. In conclusion, the results suggest a robust description of texture features with the proposed approach, which we then apply to remote sensing benchmark datasets.

\subsection{Remote Sensing Scene Classification} We evaluate our approach by performing experiments on four challenging remote sensing scene classification datasets: UC-Merced, WHU-RS19, RSSCN7 and the recently introduced AID.

\begin{figure*}[t]
\centerline{\includegraphics[width=16.7cm]{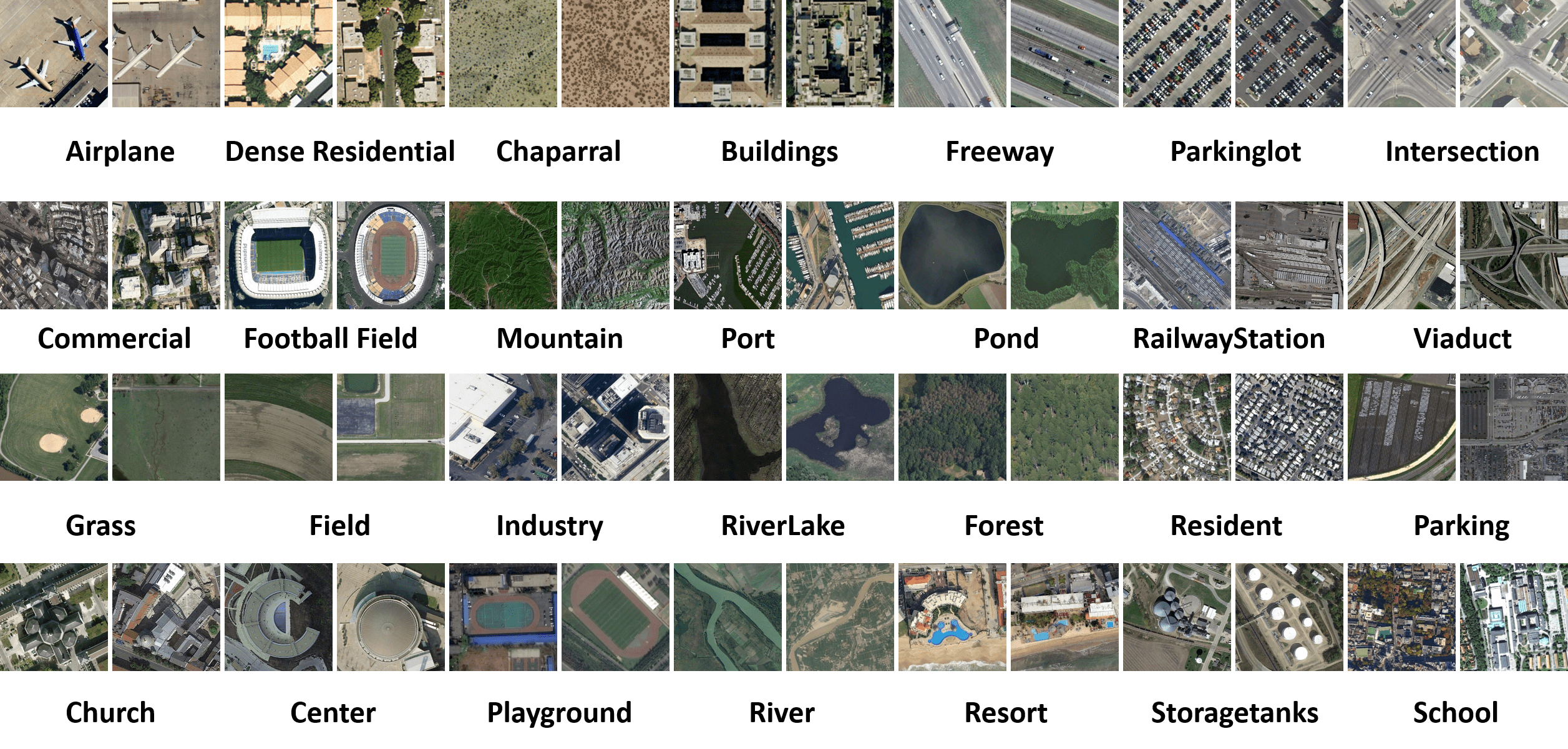}}
\caption{Example images from the four remote sensing scene classification datasets from top to bottom: UC-Merced, WHU-RS19, RSSCN7 and the recently introduced AID.  }
\label{RS_dataset}
\end{figure*}

\noindent \textbf{UC-Merced} is a publicly available dataset~\cite{Yang10j} consisting of 2100 aerial scene images with pixel resolution of one foot, downloaded from the United States Geological Survey (USGS) National Map. The images were downloaded from 20 regions across the USA: Buffalo, Boston, Birmingham, Columbus, Dallas, Houston, Harrisburg, Jacksonville, Las Vegas, Los Angeles, Miami, New York, Napa, Reno, San Diego, Santa Barbara, Seattle, Tampa, Tucson, and Ventura. The images in the dataset are cropped into 256 $\times$ 256 pixels, equally divided into 21 classes: agriculture, airplane, baseball diamond, beach, buildings, chaparral (shrubland / heathland), dense residential, forest, freeway, golf course, harbor, intersection, medium density residential, mobile home park, overpass, parking lot, river, runway, sparse residential, storage tanks, and tennis courts. The dataset is challenging with a variety of spatial land-use patterns with a significant overlap among several categories, such as medium residential, sparse residential and dense residential. These overlapping categories only differ in the density of structures.

\noindent \textbf{WHU-RS19} is a publicly available dataset~\cite{Sheng12j} consisting of 950 high spatial resolution aerial images collected from Google Earth imagery. The images in the dataset are of size 600 $\times$ 600 pixels, 50 samples per category and equally divided into 19 scene classes: airport, beach, bridge, river, forest, meadow, pond, parking, port, viaduct, residential area, industrial area, commercial area, desert, farmland, football field, mountain, park, and railway station. The dataset is challenging since images within each scene class are collected from different regions around the world with scale variations and different lighting conditions.

\noindent \textbf{RSSCN7} is a publicly available dataset~\cite{Qin15j}, released in 2015, consisting of 2800 aerial scene images. The images are divided into 7 scene classes: grassland forest, farmland, parking lot, residential region, industrial region, river, and lake. Each scene class comprises of 400 images, where each image has a size of 400 $\times$ 400 pixels. The dataset is challenging since images in each category are sampled at four different scales with different imaging angles.

\noindent \textbf{AID} is a recently introduced publicly available large-scale aerial image dataset~\cite{Xia177}. The dataset consists of 10000 images and 30 aerial scene categories: airport, bare land, baseball field, beach, bridge, center, church, commercial, dense residential, desert, farmland, forest, industrial, meadow, medium residential, mountain, park, parking, playground, pond, port, railway station, resort, river, school, sparse residential, square, stadium, storage tanks and viaduct. Unlike other aerial scene datasets, such as the UC-Merced dataset, the images in the AID dataset are collected from Google Earth imagery using different remote imaging sensors. The dataset is challenging since images in each scene category are collected from different countries around the world including China, USA, UK, France, Italy, and Germany. Further, the images are collected under varying imaging conditions (time and seasons), thereby further complicating the task of aerial scene classification. Figure~\ref{RS_dataset} shows example images from the four remote sensing scene classification datasets.

\noindent \textbf{Experimental Setup:} We follow the standard protocol~\cite{Xia177} to evaluate our approach on benchmark datasets. The performance is measured in terms of mean classification accuracy over all scene categories in a dataset. The classification accuracy is computed as $\frac{S_{p}}{S_{t}}$, where $S_{p}$ is the number of correct predictions (images) in the test set and $S_{t}$ is the total number of samples (images) in the test set. To compute the accuracy, each dataset is randomly split into training and test sets for evaluation. The evaluation procedure is then repeated ten times for a reliable performance comparison. The final results are reported as the mean and standard deviation over the ten runs. Following~\cite{Xia177}, in case of UC-Merced dataset, the ratio of training to test images was set to 50:50 and 80:20 respectively, with the images randomly selected for each category. In the case of WHU-RS19, the ratio of training to test samples was set to 40:60 and 60:40 respectively. In the case of RSSCN7 and AID datasets, the ratio of the training set was fixed to $20\%$ and $50\%$ per class respectively. As in texture recognition (section~\ref{sec:texture_exp_subsection}), we use 4096-dimensional activations from the FC7 (second last) layer as image features, where the resulting image features are $\mathrm{L_{2}}$-normalised and input to a linear SVM classifier. Consequently, we fine-tuned both our late fusion based approach and the standard RGB ResNet architecture to perform end-to-end remote sensing scene classification. For fine-tuning ResNet models, we used the same parameter settings as in texture recognition experiments.

\begin{table*}\resizebox{16.5cm}{!}{
\centering
\begin{tabular}{|c|c|c|c|c|c|c|c|c|c|}
\hline
 \textbf{Method}& \textbf{Architecture} & \textbf{UC-Merced (50$\%$)} & \textbf{UC-Merced (80$\%$)} & \textbf{WHU-RS19 (40$\%$)}& \textbf{WHU-RS19 (60$\%$)} & \textbf{RSSCN7 (20$\%$)} & \textbf{RSSCN7 (50$\%$)} & \textbf{AID (20$\%$)} & \textbf{AID (50$\%$)}  \\ \hline
Standard RGB & VGG-M & 94.13 $\pm 0.38$ & 95.40 $\pm 0.91$ & 96.01 $\pm 0.54$ & 96.57 $\pm 0.87$ & 86.0 $\pm 0.63$ & 88.8 $\pm 0.55$ & 87.70 $\pm 0.33$ & 90.29 $\pm 0.37$  \\ \hline

TEX-Net Standard & VGG-M & 91.25 $\pm 0.58$ & 92.91 $\pm 0.88$ & 92.41 $\pm 0.76$ & 94.53 $\pm 0.77$ & 83.64 $\pm 0.68$ & 86.30 $\pm 0.75$ & 82.0 $\pm 0.23$ & 85.25 $\pm 0.45$  \\ \hline
TEX-Net-EF-6ch & VGG-M & 94.36 $\pm 0.90$ & 95.27 $\pm 0.96$ & 94.71 $\pm 0.77$ & 96.0 $\pm 0.74$ & 85.65 $\pm 0.79$ & 88.70 $\pm 0.78$ & 86.84 $\pm 0.34$ & 89.68 $\pm 0.19$  \\ \hline
TEX-Net-EF-4ch & VGG-M & 94.22 $\pm 0.5$ & 95.31 $\pm 0.69$ & 95.78 $\pm 0.87$ & 96.40 $\pm 0.81$ & 86.77 $\pm 0.76$ & 89.61 $\pm 0.54$ & 87.32 $\pm 0.37$ & 90.0 $\pm 0.33$  \\ \hline
TEX-Net-LF & VGG-M & 95.89 $\pm 0.37$ & 96.62 $\pm 0.49$& 97.61 $\pm 0.36$ & 98.0 $\pm 0.52$ & 88.61 $\pm 0.46$ & 91.25 $\pm 0.58$ & 90.87 $\pm 0.11$ & 92.96 $\pm 0.18$ \\
\hline \hline
Standard RGB & ResNet & 96.22 $\pm 0.38$ & 96.80 $\pm 0.51$& 97.83 $\pm 0.38$ & 98.24 $\pm 0.53$ & 90.23 $\pm 0.43$ & 93.12 $\pm 0.55$ & 92.33 $\pm 0.13$ & 94.91 $\pm 0.19$ \\ \hline
TEX-Net-LF & ResNet & \textbf{96.91} $\pm \textbf{0.36}$ & \textbf{97.72} $\pm \textbf{0.54}$& \textbf{98.48} $\pm \textbf{0.37}$ & \textbf{98.88} $\pm \textbf{0.49}$ & \textbf{92.45} $\pm \textbf{0.45}$ & \textbf{94.0} $\pm \textbf{0.57}$ & \textbf{93.81} $\pm \textbf{0.12}$ & \textbf{95.73} $\pm \textbf{0.16}$ \\
\hline
\end{tabular}}
\caption{Baseline comparison of our Tex-Net models (overall accuracy (OA) in $\%$) with the standard RGB network on UC-Merced, WHU-RS19, RSSCN7 and AID datasets. Our late fusion based two-stream deep ResNet architecture \emph{always} outperforms the standard baseline RGB deep ResNet. }
\label{baseline_RS_results}
\end{table*}

\subsection{Baseline Comparison}
\begin{figure*}[t]
\begin{center}
\includegraphics[width=20.7cm,trim=15 80 0 0,clip=true]{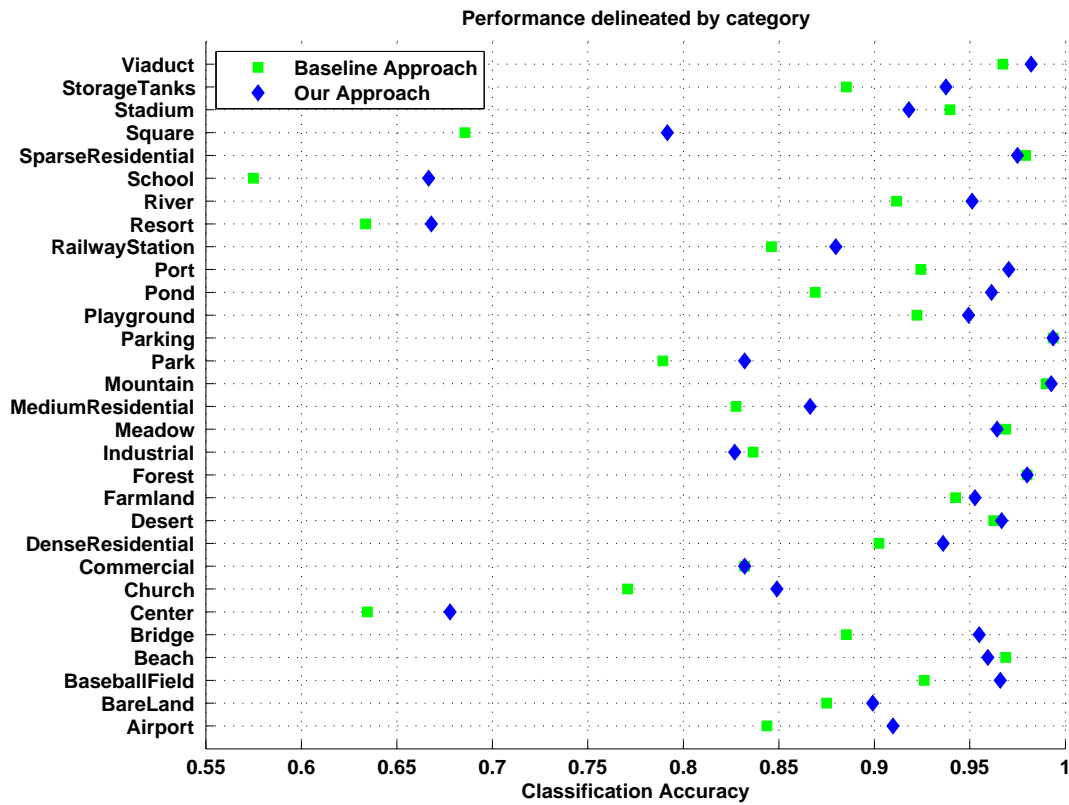}

\end{center}\vspace{-4.0cm}
\caption{Per-category performance comparison of our approach compared to the baseline RGB deep network on the AID dataset. Both the networks are based on the VGG-M architecture. Our approach improves the classification performance on most scene categories.}
\label{fig:per_class_RS}
\end{figure*}

Table~\ref{baseline_RS_results} shows the baseline comparison on four remote sensing scene classification datasets. In case of VGG-M architecture, the two early fusion based two-stream architectures provide slightly inferior performance compared to the baseline RGB network. As in texture recognition, the best results are obtained when using our late fusion based two-stream deep architecture approach, providing consistent improvements over the baseline standard RGB deep network for both VGG-M and ResNet architectures. A large gain in classification accuracy is achieved on the RSSCN7 and the large scale AID datasets. The RSSCN7 dataset comprises of several natural scene categories, such as grassland forest and farmland where texture features provide valuable complementary information to color features when other spectral channels besides RGB (like Near-Infrared) are not available. Similarly, the recently introduced large scale AID dataset consists of both natural scene types (farmland and forest) and man made scene categories (medium residential, sparse residential and school). Our late fusion approach achieves favorable results compared to the baseline RGB deep network. Figure~\ref{fig:per_class_RS} shows per-class classification performance comparison of our late fusion approach compared to the baseline RGB deep network, when using the VGG-M architecture. Our approach provides consistent improvement in performance on most scene categories.

In the seminal work of~\cite{Xia177}, it was shown that among different mid-level methods, the SIFT descriptors with the Improved Fisher Vector (IFK-SIFT) encoding provide improved results for remote sensing scene classification. Table~\ref{mid_high_RS_results} shows the comparison of our late fusion two-stream deep ResNet architecture with the best mid-level method: IFK-SIFT and several existing high-level deep methods: the shallow CaffeNet and the very deep VGG-VD-16 and GoogleNet. All the baseline results are taken from ~\cite{Xia177}.The high-level deep feature approaches obtain consistently improved performance compared to the best mid-level method IFK-SIFT. Despite having only 8 layers, CaffeNet achieves competitive performance compared to very deep VGG-VD-16 and GoogleNet. Our late fusion based two-stream ResNet architecture provides consistent gain in performance compared to the existing high-level deep methods on all four datasets. In particular, a large gain in performance is achieved on the RSSCN7 and AID datasets. On the RSSCN7 dataset (20:80 training and test set ratio), the best mid-level method (IFK-SIFT) yields a mean recognition rate of $81.08\%$. The existing high-level deep methods: CaffeNet, VGG-VD-16 and GoogleNet provide mean classification scores of $85.57\%$, $83.98\%$ and $82.55\%$ respectively. Our approach achieves a mean classification rate of $92.45\%$ outperforming best existing deep feature methods. A similar gain of $5.8\%$ in mean accuracy is achieved, compared to best existing method, with 50:50 training and test set ratio on this dataset. On the recently introduced AID dataset (20:80 training and test set ratio), the best mid-level method (IFK-SIFT) provides a mean recognition rate of $71.92\%$. The existing high-level deep methods: CaffeNet, VGG-VD-16 and GoogleNet provide mean recognition rate of $86.86\%$, $86.59\%$ and $83.44\%$ respectively. Our approach provides superior performance compared to existing methods. Furthermore, a gain of $6.1\%$ is obtained compared to the best existing deep feature method, with the 50:50 training and test set ratio on this dataset.

\begin{table*}\resizebox{16.5cm}{!}{
\centering
\begin{tabular}{|c|c|c|c|c|c|c|c|c|}
\hline

 \textbf{Method}& \textbf{UC-Merced (50$\%$)} & \textbf{UC-Merced (80$\%$)} & \textbf{WHU-RS19 (40$\%$)}& \textbf{WHU-RS19 (60$\%$)} & \textbf{RSSCN7 (20$\%$)} & \textbf{RSSCN7 (50$\%$)} & \textbf{AID (20$\%$)} & \textbf{AID (50$\%$)}  \\ \hline
IFK-SIFT & 78.74 $\pm 1.65$ & 83.02 $\pm 2.19$ & 83.35 $\pm 1.19$ & 87.42 $\pm 1.59$ & 81.08 $\pm 1.21$ & 85.09 $\pm 0.93$ & 71.92 $\pm 0.41$ & 78.99 $\pm 0.48$  \\ \hline
CaffeNet & 93.98 $\pm 0.67$ & 95.02 $\pm 0.81$ & 95.11 $\pm 1.20$ & 96.24 $\pm 0.56$ & 85.57 $\pm 0.95$ & 88.25 $\pm 0.62$ & 86.86 $\pm 0.47$ & 89.53 $\pm 0.31$  \\ \hline
 VGG-VD-16 & 94.14 $\pm 0.69$ & 95.21 $\pm 1.20$ & 95.44 $\pm 0.60$ & 96.05 $\pm 0.91$ & 83.98 $\pm 0.87$ & 87.18 $\pm 0.94$ & 86.59 $\pm 0.29$ & 89.64 $\pm 0.36$  \\ \hline
GoogleNet & 92.70 $\pm 0.60$ & 94.31 $\pm 0.89$ & 93.12 $\pm 0.82$ & 94.71 $\pm 1.33$ & 82.55 $\pm 1.11$ & 85.84 $\pm 0.92$ & 83.44 $\pm 0.40$ & 86.39 $\pm 0.55$  \\ \hline
\hline
Ours & \textbf{96.91} $\pm \textbf{0.36}$ & \textbf{97.72} $\pm \textbf{0.54}$& \textbf{98.48} $\pm \textbf{0.37}$ & \textbf{98.88} $\pm \textbf{0.49}$ & \textbf{92.45} $\pm \textbf{0.45}$ & \textbf{94.0} $\pm \textbf{0.57}$ & \textbf{93.81} $\pm \textbf{0.12}$ & \textbf{95.73} $\pm \textbf{0.16}$  \\ \hline
\end{tabular}}
\caption{Comparison of our late fusion ResNet based approach (overall accuracy (OA) in $\%$) with the best mid-level method: SIFT descriptors with Improved Fisher Vector (IFK-SIFT) encoding and the existing high-level deep methods: CafeeNet, VGG-VD-16 and GoogleNet on UC-Merced, WHU-RS19, RSSCN7 and AID datasets. Our approach provides consistently improved accuracy compared to both mid-level method and high-level deep methods on all datasets.}
\label{mid_high_RS_results}
\end{table*}

\begin{table*}[t!]
\resizebox{16.8cm}{!}{
\begin{tabular}{|c||c|c|c|c|}
\hline

\textbf{Method} &  \textbf{UC-Merced} & \textbf{WHU-RS19} & \textbf{RSSCN7} & \textbf{AID} \\ \hline

BOVW + spatial co-occurrence kernel~\cite{Yang10j}  & 77.70 & - & - & -  \\
Color Gabor~\cite{Yang10j}  & 80.50 & - & - & -  \\
SPCK + SPM~\cite{Yang11j}  & 77.40 & - & - & -  \\
Structural texture similarity~\cite{Risojevic11j}  & 86.0 & - & - & -  \\
Wavelet BOVW~\cite{Lijun14j}  & 87.40 & - & - & -  \\
Unsupervised feature learning~\cite{Cheriyadat14j}  & 81.10 & - & - & -  \\
Saliency-guided feature learning~\cite{Zhang15jj}  & 82.70 & - & - & -  \\
Concentric circle-structured BOVW~\cite{Lijun14jj}  & 86.60 & - & - & -  \\
Multifeature concatenation~\cite{Shao13j}  & 89.50 & - & - & -  \\
Pyramid-of-spatial-relations~\cite{Shizhi15jk}  &89.10& - & - & -  \\
CLBP~\cite{Chen16jk}  & 85.50 & - & - & -  \\
MS-CLBP~\cite{Chen16jk}  & 90.60 & - & - & -  \\
HHCV~\cite{Hang16jk}  & 91.80 & - & 86.40 & -  \\
DBN based feature selection~\cite{Qin15j}  & - & - & 77.0 & -  \\
Dirichlet~\cite{Kobayashi14j}  & 92.80 & - & - & -  \\
VLAT~\cite{Negrel14j}  & 94.30 & - & - & -  \\
Deep CNN Transfer (Scenario I: FC features)~\cite{Fan15jk}  & 96.88 & 96.71 & - & -  \\
Deep CNN Transfer (Scenario II: Conv features)~\cite{Fan15jk}  & 96.90 & \textbf{98.64} & - & -  \\
Deep Filter Banks~\cite{Hang16jjk}  & 92.70 & - & 90.40 & -  \\
Class-Specific Codebook + Two-Step Classification~\cite{Yan17jk}  & 93.80 & 93.70 & - & -  \\
CaffeNet~\cite{Xia177}  & 95.02 & 94.80 & 88.25 & 89.53  \\
VGG-VD-16~\cite{Xia177}  & 95.21 & 95.10 & 87.18 & 89.64  \\
GoogleNet~\cite{Xia177}  & 94.31 & 92.92 & 85.84 & 86.39  \\
\hline
This paper  & \textbf{97.72} & 98.20 & \textbf{94.0} & \textbf{95.70}  \\
\hline
\end{tabular}}\caption{Comparison (overall accuracy in \%) with the state-of-the-art approaches. Our approach provides a consistent improvement over the state-of-the-art on three datasets. Most notably a significant gain of 6.1\% is obtained, compared to the state-of-the-art, on the large scale AID dataset. Note that on the WHU-RS19 dataset, Deep CNN Transfer (Scenario II)~\cite{Fan15jk} achieves $98.64\%$ by employing VLAD encoding on the Conv layer features from the VGG-VD16. On the other hand, we do not employ any encoding scheme with the deep network.}
\label{soa_RS_tab}

\end{table*}

\subsection{State-of-the-art Comparison}
Finally, we provide a comparison with the state-of-the-art approaches in literature. Our final image representation is late fusion two-stream ResNet architecture. Table~\ref{soa_RS_tab} shows the comparison with the state-of-the-art methods in literature. We follow the same sampling setting as~\cite{Yang10j,Sheng12j,Fan15jk} for fair comparisons, by taking 80 samples per class for training in case of the UC-Merced and 30 samples per class for training in case of the WHU-RS19 dataset. In case of the RSSCN7 and AID datasets, we use 50 training samples per class for training. On the UC-Merced dataset, the approach of~\cite{Yang10j} integrating the spatial co-occurrence kernel within the bag-of-visual-words (BOVW) framework achieves a mean recognition rate of $77.7\%$. They also investigate integrating color information within Gabor features leading to a mean accuracy of $80.5\%$. The work of~\cite{Yang11j} obtains a classification accuracy of $77.4\%$ with a spatial pyramid co-occurrence based image representation that accounts for both photometric and geometric aspects of an image. Several approaches~\cite{Risojevic11j,Chen16jk,Lijun14j} aim to exploit texture information. Among these approaches, the multi-scale completed LBP feature provides superior performance with a mean recognition rate of $90.6\%$. A considerable gain in performance on this dataset can be observed with the use of deep feature based methods. The deep filter banks based approach of~\cite{Hang16jjk} achieves an accuracy of $92.7\%$. Transferring deep CNN features from the FC layer of the deep network (Deep CNN Transfer Scenario I: FC features)~\cite{Fan15jk} obtains a mean classification accuracy of $96.88\%$. Transferring deep CNNs from the Convolutional layers of the deep network encoded with the VLAD scheme (Scenario II: Conv features) achieves a recognition rate of $96.90\%$. Our approach achieves improved results ($97.72\%$) on this dataset.

On the WHU-RS19 dataset, the recently introduced improved class-specific codebook using kernel collaborative representation based classification framework~\cite{Yan17jk} achieves a mean accuracy of $93.7\%$. The CaffeNet and the very deep VGG-VD-16 and GoogleNet provide mean recognition rates of $94.8\%$, $95.1\%$ and $92.9\%$ respectively. Transferring deep CNN features from the FC layer of the deep network~\cite{Fan15jk} obtains a mean classification accuracy of $96.71\%$. Our approach achieves favorable results compared to existing methods. On this dataset, the best results ($98.6\%$) are obtained when transferring deep CNNs from the Convolutional layers of the deep network encoded using the VLAD scheme. It is worthy to mention that our approach is complementary to (Scenario II: Conv features) method~\cite{Fan15jk} and combining the two approaches can be expected to provide further gain in the classification performance.

On the RSSCN7 dataset, the deep learning based feature selection approach (DBN)~\cite{Qin15j} achieves  a mean recognition rate of $77.0\%$. The hierarchical coding vectors based classification approach~\cite{Hang16jk} achieves a classification result of $86.4\%$. The deep filter banks approach~\cite{Hang16jjk} provides a classification performance of $90.4\%$. Our approach outperforms the best existing method (deep filter banks) with a mean classification accuracy of $94.0\%$. Finally, on the recently introduced AID dataset, the CaffeNet and the very deep VGG-VD-16 and GoogleNet methods provide mean recognition rates of $89.5\%$, $89.6\%$ and $86.4\%$ respectively. Our approach achieves the best results on this dataset with a mean classification accuracy of $95.7\%$.

\section{Conclusions}
\label{conclusions}

In this paper, we address the problem
of learning robust texture description within deep learning
architectures for texture recognition and remote sensing
scene classification. We design deep models by constructing a two-stream
deep architecture where texture coded mapped
images are used as a second stream and fuse it with
the standard RGB stream. Furthermore, we investigate two fusion strategies, early and late
fusion, to combine RGB and texture streams in our two-stream
deep architecture. Experiments are conducted on several benchmark texture recognition and remote sensing scene classification datasets. Our results clearly demonstrate that the proposed late fusion two-stream deep architecture always
improves the overall performance compared to the standard
RGB stream deep network architecture for both recognition tasks. Further, our final combination leads to improved results compared to
the state-of-the-art for remote sensing scene classification. In this paper, we investigate Local Binary Patterns (LBP) encoded CNNs and different deep network fusion architectures for texture recognition and remote sensing scene classification. Future work involves investigating alternative texture description techniques and fusion strategies for texture coded deep CNNs. Another future direction is to include training and testing the proposed approach on actual full-sized satellite images containing all available spectral bands besides RGB (e.g. Near Infrared).

\section*{Acknowledgements}
\noindent This work has been funded by the Spanish project TIN2016-79717-R, the CHISTERA project M2CR (PCIN2015-251), SSF through a grant for the project SymbiCloud, VR starting grant (2016-05543), through the Strategic Area for ICT research ELLIIT, CENIIT grant (18.14), the project AIROBEST (317387, 317388) funded by the Academy of Finland, the project {MegaM\@rt2}, funded by the Electronic Component Systems for European Leadership (ECSEL) Joint Undertaking (grant agreement No. 737494) of the Horizon 2020 European Union funding programme. We acknowledge the computational resources provided by the Aalto Science-IT project and CSC IT Center for Science, Finland. We also acknowledge the computational support from Nvidia and the NSC.

{\small
\bibliographystyle{elsarticle-num}
\bibliography{egbib_new1}
}

\end{document}